# An Efficient Automatic Attendance System Using Fingerprint Reconstruction Technique

Josphineleela.R
Research scholar
Department of Computer Science and Engineering
Sathyabamauniversity
Chennai,India
ilanleela@yahoo.com

Dr.M.Ramakrishnan
Professor/HOD-IT
Velammal Engineering College
Chennai,India
ramkrishod@gmail.com

*Abstract*— **Biometric time and attendance system is one of the most successful applications of biometric technology. One of the main advantage of a biometric time and attendance system is it avoids "buddy-punching". Buddy punching was a major loophole which will be exploiting in the traditional time attendance systems. Fingerprint recognition is an established field today, but still identifying individual from a set of enrolled fingerprints is a time taking process. Most fingerprint-based biometric systems store the minutiae template of a user in the database. It has been traditionally assumed that the minutiae template of a user does not reveal any information about the original fingerprint. This belief has now been shown to be false; several algorithms have been proposed that can reconstruct fingerprint images from minutiae templates. In this paper, a novel fingerprint reconstruction algorithm is proposed to reconstruct the phase image, which is then converted into the grayscale image. The proposed reconstruction algorithm reconstructs the phase image from minutiae. The proposed reconstruction algorithm is used to automate the whole process of taking attendance, manually which is a laborious and troublesome work and waste a lot of time, with its managing and maintaining the records for a period of time is also a burdensome task. The proposed reconstruction algorithm has been evaluated with respect to the success rates of type-I attack (match the reconstructed fingerprint against the original fingerprint) and type-II attack (match the reconstructed fingerprint against different impressions of the original fingerprint) using a commercial fingerprint recognition system. Given the reconstructed image from our algorithm, we show that both types of attacks can be effectively launched against a fingerprint recognition system.**

*Keywords—Fingerprint Reconstruction, attendance management system, Minutiae Extraction*

## I. INTRODUCTION (HEADING 1)

Fingerprint reconstruction is one of the most well-known and publicized biometrics. Because of their uniqueness and consistency over time, fingerprints have been used for identification over a century, more recently becoming automated due to advancements in computed capabilities. Fingerprint reconstruction is popular because of the inherent ease of acquisition, the numerous sources (e.g. ten fingers) available for collection, and their established use and collections by law enforcement and immigration.

Minutiae-based fingerprint matching algorithm [1] has been proposed to solve two problems: correspondence and similarity computation. For the correspondence problem, use an alignment-based greedy matching algorithm to establish the correspondences between minutiae.

Cryptographic techniques are being widely used for ensuring the secrecy and authenticity of information. Although several cryptosystems have proven security guarantees (e.g., AES and RSA), the security relies on the assumption that the cryptographic keys are known only to the legitimate user. Maintaining the secrecy of keys is one of the main challenges in practical cryptosystems. However, passwords can be easily lost, stolen, forgotten, or guessed using social engineering and dictionary attacks. Limitations of password-based authentication can be alleviated by using stronger authentication schemes, such as biometrics. Biometric systems establish the identity of a person based on his or her anatomical or behavioral traits, such as face, fingerprint, iris, voice, etc. Biometric authentication is more reliable than password-based authentication because biometric traits cannot be lost or forgotten and it is difficult to share or forge these traits. Hence, biometric systems offer a natural and reliable solution to the problem of user authentication in cryptosystems.

Reliable information security mechanisms are required to combat the rising magnitude of identity theft in our society. While cryptography is a powerful tool to achieve information security, one of the main challenges in cryptosystems is to maintain the secrecy of the cryptographic keys. Though biometric authentication can be used to ensure that only the legitimate user has access to the secret keys, a biometric system itself is vulnerable to a number of threats.

A critical issue in biometric systems is to protect the template of a user which is typically stored in a database or a smart card. The fuzzy vault construct is a biometric cryptosystem that secures both the secret key and the biometric template by binding them within a cryptographic framework. In [2], fuzzy vault scheme has been proposed based on fingerprint minutiae. Since the fuzzy vault stores only a transformed version of the template, aligning the query fingerprint with the template is a challenging task. We extract high curvature points derived from the fingerprint orientation field and use them as helper data to align the template and query minutiae. The helper data itself do not leak any information about the minutiae template, yet contain sufficient information to align the template and query fingerprints accurately. Further, we apply a minutiae matcher during decoding to account for nonlinear distortion and this leads to significant improvement in the genuine accept rate. The performance improvement can be achieved by using multiple fingerprint impressions during enrollment and verification.

Because of the stability and uniqueness, fingerprint is widely used in biometric identification. The matching method is one of the most crucial technologies in the Automated Fingerprint Identification System (AFIS). Whether two fingerprints are matched relies on the similarity measure between the effective features of them. There are mainly two kinds of features used in fingerprint matching: local features and global features. Two most prominent local ridge characteristics, called minutiae, are ridge ending and ridge bifurcation. Minutiae are the most widely used features in the matching process.

The performance of Automated Fingerprint Identification System (AFIS) is highly defined by the similarity of effective features in fingerprints. Minutia is one of the most widely used local features in fingerprint matching. In [3], proposes two global statistical features of fingerprint image, including the mean ridge width and the normalized quality estimation of the whole image, and proposed a novel fingerprint matching algorithm based on minutiae sets combined with the global statistical features. The algorithm proposed in this paper has the advantage of both local and global features in fingerprint matching. It can improve the accuracy of similarity measure without increasing of time and memory consuming.

The non–linear distortion in the fingerprint images makes it very difficult to handle matching as it changes the geometrical position of the minutiae points. The regions, that are affected, shift the geometry of the minutiae and hence pose a potential threat to acceptance of a genuine match. The distortion is due to the pressure applied on the scanner, the static friction, the skin moisture, elasticity, and rotational effects, which occur during the acquisition. The level of distortion increases from the center towards the outer regions. The existing approaches for fingerprint matching are: minutiae–based, and correlation-based. The former has several advantages over the latter such as lower time complexity, better space complexity, less requirement of hardware etc.

The uniqueness of a fingerprint is due to unique pattern shown by the locations of the minutiae points– irregularities of a fingerprint–ridge endings, and bifurcations. A novel minutiae-based approach [4], has been proposed to match fingerprint images using similar structures. Distortion poses serious threats through altered geometry, increases false minutiae, and hence makes it very difficult to find a perfect match. This algorithm divides fingerprint images into two concentric circular regions – inner and outer – based on the degree of distortion. The algorithm assigns weight ages for a minutiae–pair match based on the region in which the pair exists. The algorithm has two stages. In the first stage, the minutiae points are extracted, and in the second stage, the aligning and the matching of the fingerprint images are done. The algorithm is designed to reduce time taken in aligning, immediately after the calculation of the binary image.

Recent advances in automated fingerprint identification technology, coupled with the growing need for reliable person identification have resulted in an increased use of fingerprints in both government and civilian applications such as border control, employment background checks, and secure facility access. In [5], Quadratic differentials naturally define analytic orientation fields on planar surfaces. This method proposed model orientation fields of fingerprints by specifying quadratic differentials which is used for reliable person identification. Models for all fingerprint classes such as arches, loops and whorls are laid out. These models are parameterized by few, geometrically interpretable parameters which are invariant under Euclidean motions. Potential applications of these models are the use of their parameters as indices of large fingerprint databases, as well as the definition of intrinsic coordinates for single fingerprint images. The accuracy of models is still challenging task for arches.

General characteristics of the fingerprint emerge as the skin on the fingertip begins to differentiate. Fingerprint recognition systems have the advantages of both ease of use and low cost. Because among various biometric identifiers, such as face, signature, and voice, the fingerprint has one of the highest levels of distinctiveness and performance and it is the most commonly used biometric modality. Haiyun Xu et. al., [6], proposed a novel method to represent minutiae set as a fixed-length feature vector, which is invariant to translation, and in which rotation and scaling become translations, so that they can be easily compensated for recognition. These characteristics enable the combination of fingerprint recognition systems with template protection schemes that require a fixed-length feature vector. This method introduces the concept of algorithms for two representation methods: the location-based spectral minutiae representation and the orientation-based spectral minutiae representation. Both algorithms are evaluated using two correlation-based spectral minutiae matching algorithms. The performance can be improved by using a fusion scheme and singular points. The spectral minutiae representation overcomes the drawbacks of the minutiae sets, thus broadening the application of minutiae-based algorithms. The minutiae extractor is not reliable it affects the efficiency of spectral minutiae representation.

Automated Fingerprint Identification Systems (AFISs) have played an important role in many forensics and civilian applications. There are two main types of searches in forensics AFIS: ten print search and latent search. In ten print search, the rolled or plain fingerprints of the 10 fingers of a subject are searched against the fingerprint database of known persons. In latent search, a latent print developed from a crime scene is searched against the fingerprint database of known persons. Latent Fingerprint matching [7], propose a system for matching latent fingerprints found at crime scenes to rolled fingerprints enrolled in law enforcement databases which overcomes the difficulties in poor quality of ridge impressions, small finger area, and large nonlinear distortion. In addition to minutiae, extended features are also used including singularity, ridge quality map, ridge flow map, ridge wavelength map, and skeleton. In order to evaluate the relative importance of each extended feature, these features were incrementally used in the order of their cost in marking by latent experts. The matching accuracy should be improved.

Despite tremendous progress made in automatic fingerprint identification systems, matching incomplete o partial fingerprints such as latent prints remains a critical challenge today. Existing partial fingerprint algorithms concentrate on improving the accuracy of one-to-one matching based on local ridge details However, the performance of one-to-one matching depends on image quality as well as the number of high-level features detectable in the partial fingerprint segments. These ad hoc algorithms are designed on the basis of more delicate one-to-one comparisons. When used in one-to-many applications, they generally assume sequential matching or that the candidate list for such matching has already been established. However, sequential matching is not efficient for large-scale identification, which can involve thousands or millions of records in the target database, and retrieving a short and reliable list of candidates for matching is difficult in practice. An innovative method [8], propose an analytical approach for reconstructing the global topology representation from a partial fingerprint. Analytical approach solves the problem of retrieving candidate lists for matching partial fingerprints by exploiting global topological features. First, an inverse orientation model for describing the reconstruction problem is presented. Then, a general expression for all valid solutions to the inverse model is provided. This allows us to preserve data fidelity in the existing segments while exploring missing structures in the unknown parts. Further developed algorithms for estimating the missing orientation structures based on some a priori knowledge of ridge topology features are described. The statistical experiments show that the proposed model-based approach can effectively reduce the number of candidates for pair wised fingerprint matching, and thus significantly improve the system retrieval performance for partial fingerprint identification.

Fingerprint matching systems generally use four types of representation schemes: grayscale image, phase image, skeleton image, and minutiae, among which minutiae-based representation is the most widely adopted one. It has been traditionally assumed that minutiae template does not retrieve any information about original fingerprint. In [9], three levels of information about the parent fingerprint can be elicited from a given minutiae template: the orientation field, the fingerprint class, and the friction ridge structure. The orientation estimation algorithm determines the direction of local ridges using the evidence of minutiae triplets. The estimated orientation field, along with the given minutiae distribution, is then used to predict the class of the fingerprint. Finally, the ridge structure of the parent fingerprint is generated using streamlines that are based on the estimated orientation field. Line Integral Convolution is used to impart texture to the ensuing ridges, resulting in a ridge map resembling the parent fingerprint. But the visual appearance of reconstructed fingerprint is not accurate.

The location, position, as well as the type and quality of the "minutiae" are factors taken into consideration in the template creation stage. A minutiae-based template did not contain enough information to allow the reconstruction of the original fingerprint. A novel approach [10], is proposed to reconstruct fingerprint images from standard templates and examines to what extent the reconstructed images are similar to the original ones. The efficacy of the reconstruction technique has been assessed by estimating the success chances of a masquerade attack against nine different fingerprint recognition algorithms. The experimental results show that the reconstructed images are very realistic and that, although it is unlikely that they can fool a human expert, there is a high chance to deceive state-of-the-art commercial fingerprint recognition systems.

The fingerprint recognition system is used for person authentication and identification in industries and many commercial appliances. The fingerprint recognition does not have the efficiency in the case of fake fingerprints which extracts minutiae from templates. The compactness of minutiae representation has created an impression that the minutiae template does not contain sufficient information to allow the reconstruction of the original grayscale fingerprint image. In [11], a novel fingerprint reconstruction algorithm is proposed to reconstruct the phase image, which is then converted into the grayscale image. Reconstruction algorithm not only gives the whole fingerprint, but the reconstructed fingerprint contains very few spurious minutiae. A fingerprint image is represented as a phase image which consists of the continuous phase and the spiral phase. The proposed reconstruction algorithm has been evaluated with respect to the success rates of type-I attack and Type II attacks using a commercial fingerprint recognition system. Reconstruction algorithm should be modified in order to apply the important problems of latent fingerprint restoration. The proposed reconstruction algorithm is used to automate the whole process of taking attendance, manually which is a laborious and troublesome work and waste a lot of time, with its managing and maintaining the records for a period of time is also a burdensome task.

## II. ATTENDANCE MANAGEMENT SYSTEM

Attendance management system is one of the most successful applications of biometric technology. With the integration and use of biometric technology getting simpler, many institutions are venturing down the biometric road to verify the time and attendance of their students and staffs.

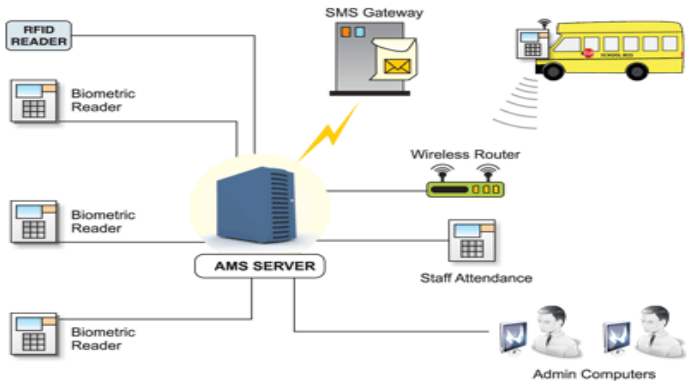

Figure 1.Attendance Management System(AMS)

## III. Related work

In this system the fingerprint is taken as an input for attendance management and it is organized into the following modules Pre-processing, Minutiae Extraction,Reconstruction,FingerprintRecognition, Report generation

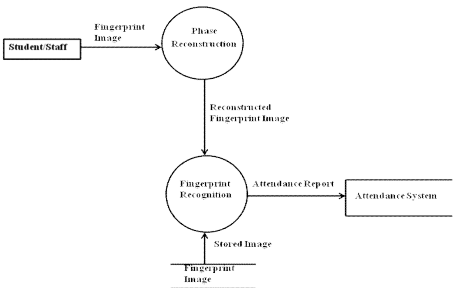

Fig 2.DFD for Attendance Management System

## SYSTEM ARCHITECTURE

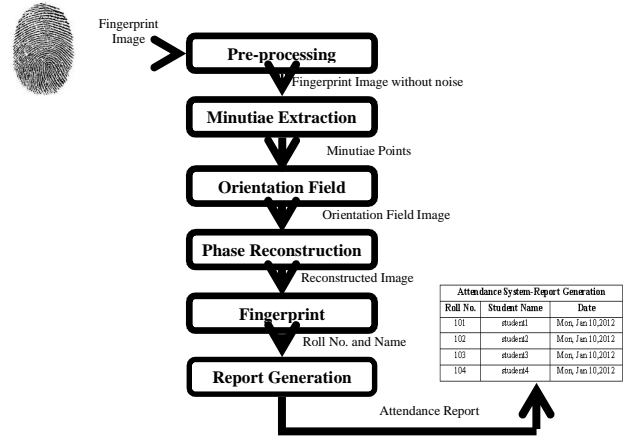

### IV. Preprocessing

There are two steps in Pre-processing

**Step 1:Segmentation**

**Step 2: Normalization**

#### 1 Segmentation

- Image segmentation separates the foreground regions and the background regions in the fingerprint image.
- Segmentation is a process by which can discard these background regions, which results in more reliable extraction of minutiae points.

#### 2 Normalization

- Normalization is a process of standardizing the intensity values in an image so that these intensity values lie within a certain desired range.
- It can be done by adjusting the range of grey-level values in the image.

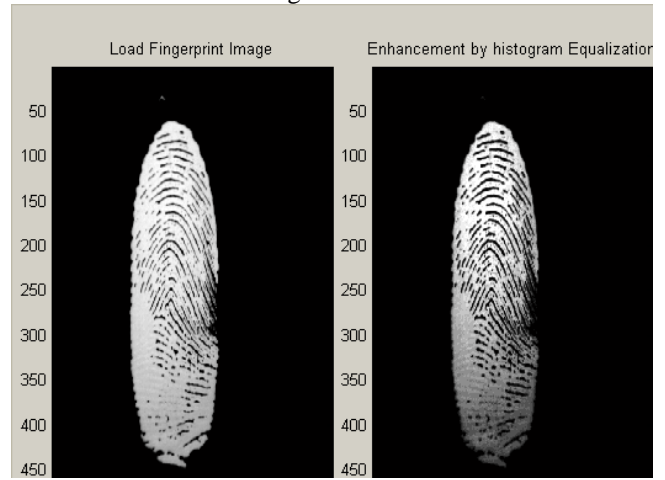

Fig 3 Result of histogram equalization (a) original image (b) After histogram equalization

## V. Minutiae Extraction

Minutiae points are extracted from composite phase image of fingerprint image which is obtained by adding spiral phase to the continuous phase.

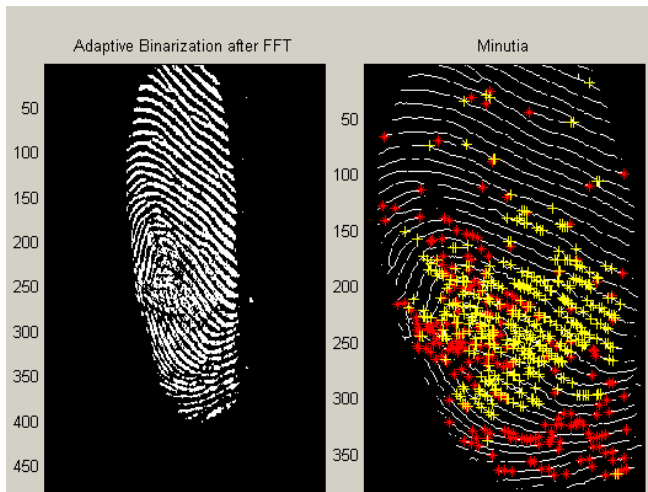

Fig 4 Result of Minutiae Extraction

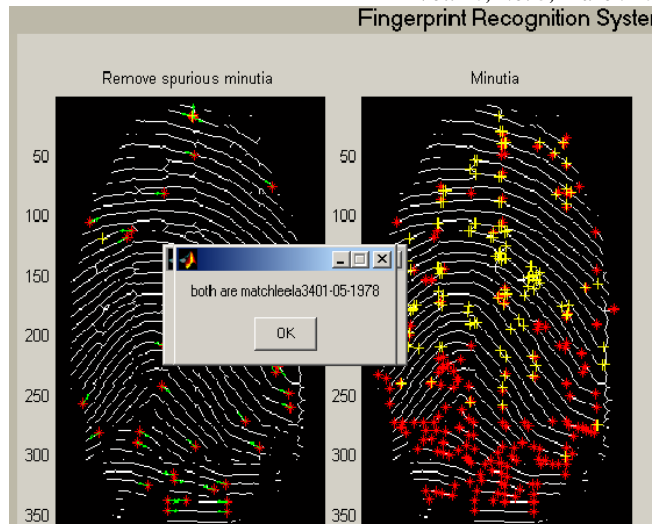

Fig 5 Result of Minutiae Matching

## VI. Fingerprint Reconstruction

There are two steps in reconstruction

**Orientation Field Reconstruction**
- An orientation field reconstruction algorithm that can work even when only one minutia is available.
- The image is divided into non overlapping blocks of 8 x 8 pixels and an orientation value is computed for each foreground block.

**Phase Reconstruction**
- The continuous phase has been reconstructed at all of the foreground blocks by estimating the phase offset value.
- The reconstructed phase image validates the minutiae points and eliminates spurious minutiae.

## VII. Fingerprint Recognition

Fingerprint is recognized if the reconstructed fingerprint matches with the original fingerprint.

## VIII. Result

The report will be generated with Roll number of the matched fingerprint and stored in an attendance system.

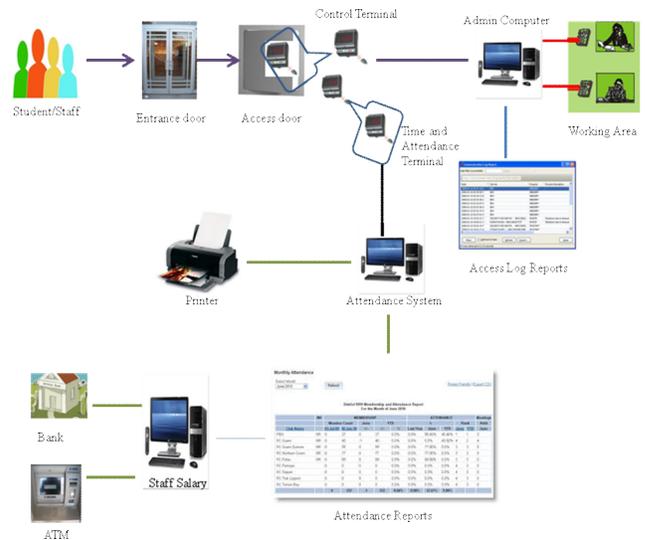

## IX. Conclusion

The proposed system will make way for perfect management of students and staff attendance and produce more accuracy

AUTHORS PROFILE

Authors Profile …

**R.Josphineleela** received the B.SC computer science degree from the Department of Computer Science Madurai KamarajUniversity. Madurai, India, in 1998 and M.C.A degree in the same university in 2001.She received the M.E.degree from the Department of Computer Science and Engineering Sathyabama University, Chennai, India, in 2007.She has published Four papers in International Level Conferences and Three papers in National level Conferences .She has published one paper in International Journal. She has 10 years teaching experience and was awarded best teacher in the year 2011 by Panimalar Engineering College,chennai. She is pursuing her PhD under the guidance of Dr.M.Ramakrishan .Her research interests are in Image processing, Pattern recognition, soft computing and artificial neural network etc.

**Dr.M.Ramakrishan** was born in 1967. He is working as a Professor and Head of PG department of Computer Science and Engineering in velammal Engineering College, Chennai. He is a guide for research scholars in many universities .His area of interest is Parallel Computing, Image Processing, Web Services and Network Security. He has 21 years of teaching experience and published 8 National and International journals and 40 National and International Conferences. He is member of ISTE and senior member of IACSIT.